\documentclass[12pt]{spieman}  
\usepackage{amsmath,amsfonts,amssymb}
\usepackage{graphicx}
\usepackage{setspace}
\usepackage{array, makecell}
\usepackage{tocloft}
\usepackage{algorithm2e}
\usepackage{subcaption}
\usepackage{xcolor}
\usepackage{lineno}
\usepackage{xcolor, soul}

\title{CARDIE:Clustering Algorithm on Relevant Descriptors for Image Enhancement}

\author[a,b]{Giulia Bonino}
\author[b]{Luca Alberto Rizzo*}
\affil[a]{Eurecom, Huawei Nice Reseach Center}
\affil[b]{Huawei Nice Research Center}

\cftpagenumbersoff{figure}
\cftpagenumbersoff{table} 
\begin{document} 
\maketitle
\newcommand{\clustname}{\textit{CARDIE} } 
\newcommand{\silscore}{$\mathrm{SC}^{'}$ } 

\begin{abstract}
{Automatic image clustering is a cornerstone of computer vision, yet its application to image enhancement remains limited by the challenge of defining meaningful clusters tailored to this task. To address this issue, we present a novel clustering algorithm} \clustname {which clusters images in an unsupervised manner according to their color and luminosity content. We also introduce a new method to quantify the impact of image enhancement algorithms on luminance distribution and local variance.  Using this method, we demonstrate that} \clustname{produces clusters more relevant to image enhancement than those obtained through semantic image attributes. Finally, we show that clusters defined by} \clustname {can be  used effectively to resample image enhancement datasets, leading to improved performances for tone mapping and denoising algorithms.  To foster} \clustname {usage and ensure the reproducibility of the analysis conducted here, we release} \clustname {code on our} \href{https://github.com/GiuliaBonino/CARDIE}{ github}.
\end{abstract}

\keywords{image processing, automatic image clustering, resampling}

{\noindent \footnotesize\textbf{*}Luca Alberto Rizzo,  \linkable{luca.alberto.rizzo@huawei.com} }

\begin{spacing}{2}   


\section{Introduction}
\label{sec:introduction}  
Traditional and CNN-based algorithms have been used for a variety of image enhancement (IE) tasks with impressive results (e.g. denoising \cite{chen2022simplebaselinesimagerestoration} or tone mapping \cite{Zhao2022, GharbiCBHD17}).
However, despite their successes, these methods remain highly sensitive to the characteristics of the training dataset, leading to a rapid decline in performance when applied to images that deviate significantly from the ``average" ones (e.g. images that are too ``bright”) \cite{hashmani2019accuracy}. 

For computer vision tasks for which classes are naturally present (e.g image classification)  this problem is often
alleviated by the use of resampling techniques such as under/over sampling \cite{Khan2023}. However, to the best of our knowledge, no standard solution has been developed for IE, partly because image classes cannot be naturally defined.

 {Numerous efforts have been made to address this issue by automatically clustering images using large neural networks trained to recognize semantic content.

One possible approach is to leverage extensive repositories of manually annotated with semantic categories,  } \cite{ zhou2017places, zhou2018semanticunderstandingscenesade20k, lin2015microsoftcococommonobjects, 5539970}  { to train CNNs to recognize image content for scene classification. Although these methods yield impressive results across diverse scenes, they are highly dependent on large pre-trained models. For this reasons, as we will show in Sec.} \ref{sec:results_resampling} {, they prove ineffective for image enhancement tasks due to significant disparities between the datasets used for semantic recognition and those employed in image enhancement applications.}

{To eliminate the need for manual dataset annotation, some authors} \cite{VanGansbeke2020, caron2021unsupervisedlearningvisualfeatures}  {proposed an approach that automatically groups images into semantically meaningful clusters by first learning feature representations and then using them as priors in a clustering framework.  For a similar reasons, other authors} \cite{Niu_2022} { propose a joint training approach that combines a feature extraction model with a clustering head, using pseudo-labeling techniques to improve clustering performance. However, regardless of the specific approach used for semantic feature representations and clustering, these methods inherently prioritize high-level content—such as objects and scenes—while overlooking crucial low-level attributes like color and luminance, making them unsuitable for IE tasks.}

For these reasons, we propose a novel  clustering approach,  \clustname  which first focuses on extracting descriptors that are meaningful for IE tasks and then clusters them in an unsupervised manner. As we will show in Sec.\ref{sec:results}, this approach not only clusters images in a more relevant way for IE but also relies solely on information that can be readily extracted from the IE dataset, removing the need of labor-intensive  manual image labelling or expensive training.

We also introduce a  methodology which quantifies the modification of image luminance distributions and average local variance between  input and output images induced by the IE algorithm. In Sec.\ref{sec:results}, we use this information, combined with cluster labels,  to assess cluster relevance by verifying if images within each cluster undergo similar IE transformations.

Finally, we exploit the clusters obtained using \clustname to resample the training dataset for tone mapping and denoising tasks, leading to substantial improvements in the performances of state-of-the-art tone mapping and denoising algorithms.

\color{black}
In summary, our paper's main contributions are:
\begin{itemize}

\item[--] introducing \clustname, a unsupervised algorithm based on color and luminance information and HDBSCAN \cite{CampelloHDBSCAN}, which clusters IE datasets in a relevant way
\item[--] introducing a  method to quantify the effect of IE algorithms for large datasets, based on the modifications of the image luminance distribution for tone mapping and average local variance for denoising 
\item[--] quantifying the relevance of two unsupervised clustering algorithms for tone mapping and denoising tasks, showing that \clustname  outperforms semantic clustering algorithms on the MIT5K and HDR+ datasets.
\item[--] employing \clustname's clusters to resample the training dataset for tone mapping and denoising, achieving better picture quality with respect to the model trained on the original dataset

\end{itemize}

To support reproducibility and facilitate the use of \clustname, all code employed in this analysis is publicly accessible on our \href{https://github.com/GiuliaBonino/CARDIE}{ github}.

\section{Proposed Method}
\label{section:method}

The approach presented in this paper centers on two main techniques: an unsupervised clustering algorithm that utilizes image color and luminance, and a framework for evaluating the impact of image enhancement algorithms by analyzing modifications in luminance distribution and average local variance between input and output images.

For future reference, we define here:
\begin{itemize}
	\item the image dataset $\mathcal{D} = \left \{ \left (x^{i}_1, x^{gt}_1   \right ), \cdots, \left(x^{i}_N, x^{gt}_N \right)    \right \}$, where $ N$ is the total number of images and the indexes $i$ and $gt$ represent respectevely the input and ground truth images 
	\item $L(x^{k}_j, p)$ is the luminance distribution evaluated on the $j$-th image of the dataset (input or ground truth one  $k = \left \{ i, gt  \right \}$) and $p$ as pixel index (see App.\ref{app:lum_hue_distr} for more details)
	\item $\mathcal{H}(x^{k}_j, \theta)$  hue distribution evaluated on the $j$-th image of the dataset (input or ground truth one  $k = \left \{ i, gt  \right \}$), with the hue angle $\theta$ expressed in radians (see App.\ref{app:lum_hue_distr} for more details) 
	\item $\mathcal{C} = \left ( \mathcal{C}_1, \cdots, \mathcal{C}_l    \right )$, the list of $l$ possible dominant colors, i.e. all the colors the algorithm uses to potentially flag an image  
\end{itemize}  

\subsection{CARDIE}
\label{subsection:clustering_algo}

\clustname is a two-step procedure, which first computes image descriptors based on  luminance and hue distributions and then leverages them to cluster images in an unsupervised way using HDBSCAN.

In this section we will focus on describing how \clustname is applied on input images $x^{i}_j$, because it represents the most interesting case for our discussion, even though  it can be similarly applied to any set of images.

\subsubsection{Image descriptors algorithm}
\label{subsection:image_descriptor}

\begin{algorithm}
\caption{Image descriptors algorithm}

\textit{Compute dataset luminance thresholds}

\textit{Subsample $K$ images from $\mathcal{D}$}

\textit{Initialize dataset luminance distribution} $\bar{L}(\mathcal{D}_K,p)$;

\For{image $x_j \in \mathcal{D}_K$}{
	compute $L(x_j,p)$;
	
	$\bar{L}(\mathcal{D}_K,p) = \bar{L}(\mathcal{D}_K,p) + L(x_j,p)$ 
}

$\bar{L}(\mathcal{D}_K,p) = \bar{L}(\mathcal{D}_K,p) / K$;

$\bar{L}^{\mathcal{D}_K}_{\mathrm{low}} = \mathrm{Perc}(\bar{L}(\mathcal{D}_K,p),20)$; 

$\bar{L}^{\mathcal{D}_K}_{\mathrm{high}} = \mathrm{Perc}(\bar{L}(\mathcal{D}_K,p),80)$;

\textit{Compute luminance and color flags for each image}

\For{image $x_j \in \mathcal{D}$}{
	\textit{Image luminosity}
	
	compute $L(x_j,p)$;
	
	compute $\bar{L}_j = \mathrm{med}(L(x_j,p))$ ;
	
	\If{$\bar{L}_j < \bar{L}_{\mathrm{low}} $}{
		flag $x_j $ as  low luminosity;
	}
	\ElseIf{$\bar{L}_j > \bar{L}_{\mathrm{high}} $}{
		flag $x_j$ as  high luminosity;
	}
	\Else{
		flag $x_j$ as  average luminosity;
	}
	
	\textit{Image dominant colors}
	
	compute $\mathcal{H}(x_j, \theta)$
	
	\For{$\mathrm{C}_m \in \mathrm{C}$}{
		
		\If{$\int_{\mathcal{C}_m - \Delta \theta / 2}^{\mathcal{C}_m + \Delta \theta / 2 }  \mathcal{H}(x_j, \theta) d \theta \geq  \frac{1}{l}$}{
			assign $\mathrm{C}_m$ as dominant color 
		
		}

	}

}

\label{algo:clustername}
\end{algorithm}

The first step of \clustname is to compute image descriptors for each $x_j^{i}$. As shown in Alg.\ref{algo:clustername}, the image descriptor algorithm is composed of 3 main phases. Firstly,  we compute the luminosity of a fictitious ``average image" $\bar{L}(\mathcal{D}_K, p)$  calculated as the per-pixel mean of a subset of the dataset
\begin{equation}
	\bar{L}(\mathcal{D}_K, p) = \frac{1}{K} \sum_{j=1}^{K} L(x_{j}, p).
	\label{eq:avg_luminosity}
\end{equation}
where $K \leq N$ is a free parameter of \clustname. Given Eq.\ref{eq:avg_luminosity}, we define the thresholds of low and high luminosity, respectively $\bar{L}^{\mathcal{D}_K}_{\mathrm{low}}$ and $\bar{L}^{\mathcal{D}_K}_{\mathrm{high}}$, as the 20th and 80th percentile of $\bar{L}(\mathcal{D}_K, p)$ (see Fig.\ref{fig:lum_group_hue_distribution}).

Once that $\bar{L}^{\mathcal{D}_K}_{\mathrm{low}}$ and $\bar{L}^{\mathcal{D}_K}_{\mathrm{high}}$ are computed, \clustname computes the luminosity level for each image  by first computing $\bar{L}_j$, the median of $\bar{L}(x_j, p)$ and then flagging the image  as low luminosity, if $\bar{L}_j < \bar{L}^{\mathcal{D}_K}_{\mathrm{low}}$, average luminosity  if $ \bar{L}^{\mathcal{D}_K}_{\mathrm{low}} \leq \bar{L}_j \leq \bar{L}^{\mathcal{D}_K}_{\mathrm{high}}$ or high luminosity if $\bar{L}_j > \bar{L}^{\mathcal{D}_K}_{\mathrm{high}} $.

To compute the color descriptors, \clustname computes $\mathcal{H}(x_j, \theta)$ of each image and then assigns $\mathcal{C}_m$ as dominant color to an image if:
\begin{equation}
	 \int_{\mathcal{C}_m - \Delta \theta / 2}^{\mathcal{C}_m + \Delta \theta / 2 }  \mathcal{H}(x_j, \theta) d \theta \geq  \frac{1}{l}
	 \label{eq:dom_colors}
\end{equation}
i.e. if more than $1/l$ of its hue distribution is contained in a $\Delta \theta$ interval around the $\mathcal{C}_m$ (see Fig.\ref{fig:lum_group_hue_distribution}).

Therefore, \clustname maps each image  into an array of (relatively few) descriptors $x_j \rightarrow \left ( d_{j,1}, \cdots, d_{j,k} \right )$ (with  $d_{j,1}=\left \{ 0,1  \right \}$),  allowing us to shift from an image dataset to a tabular one, severely reducing the complexity of the unsupervised clustering task.

\subsubsection{Unsupervised clustering}
\label{subsection:unsuper_clustering}

 \clustname {performs the unsupervised clustering task using HDBSCAN, a hierarchical density-based algorithm which demonstrated excellent performances on tabular data} \cite{CampelloHDBSCAN}  { and exhibits greater robustness against noise compared to K-means and spectral clustering methods. }. \cite{baligodugula2025unsupervisedlearningcomparativeanalysis}.

 {HDBSCAN  infers automatically the number of clusters from the data and requires the user to specify only on the minimum cluster size $m_{min}$  and distance function.}

 {Given the boolean nature of the descriptors presented in} \ref{subsection:image_descriptor},  {we fix the Jaccard index as distance function for HDBSCAN and we treat  $m_{min}$ as a hyperparameter.}

 {Moreover, to avoid using descriptors that yield little information, we exclude those whose variance computed on $\mathcal{D}$ is lower than a threshold $\sigma_d$, treated as an additional hyperparameter.}

 {Leveraging the low computational cost of HDBSCAN, we determine the optimal cluster labels $c_l$ for each image by conducting a hyperparameter search over its parameters $m_{min}$ and $\sigma_d$. This search is performed on a regular grid of approximately 200 values, with the silhouette score} \silscore {serving as the evaluation metric.} \cite{ROUSSEEUW198753}

\begin{figure*}[!t]
	\centering
	\includegraphics[scale=0.4]{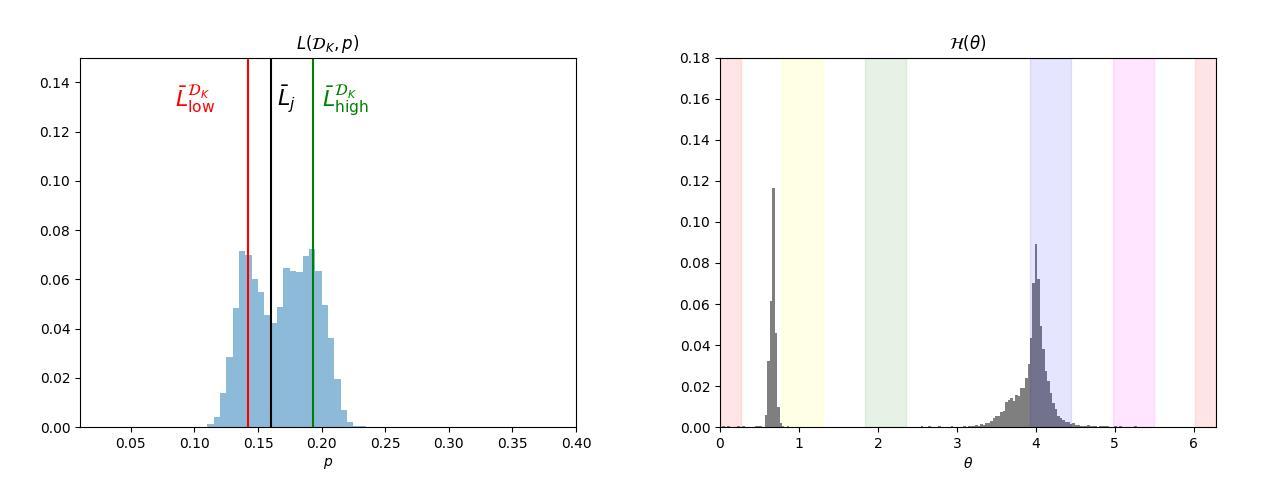}
	\caption{Examples of $L(\mathcal{D}_K,p)$ and $\mathcal{H}(\theta)$ for image labelled as \textit{AvB} (see Table \ref{clusters_tm}) by Alg. \ref{algo:clustername}}   
	\label{fig:lum_group_hue_distribution}
\end{figure*}

\subsection{Image enhancement algorithms quantification}
\label{subsection:quantify_algo}

As discussed in Sec.\ref{sec:introduction}, we  propose here a method to quantify the effect of IE algorithm $\mathcal{A}$ on the image luminance distributions and local pixel variance.

Firstly, we consider $\mathcal{A}$ (e.g. a tone mapper) as an operator which  modifies the luminance distribution of each image $x_j$ in such a way:
\begin{equation}
	L(x^{pp}_j, p) = \mathcal{A}(L(x^{i}_j), p).
	\label{eq:lum_operator}
\end{equation}
i.e. an operator that maps, for each pixel $p$, the input image $x^{i}_j$ luminance distribution to the post-processed image $x^{pp}_j$. Eq.\ref{eq:lum_operator} suggests that we can quantify the impact of $\mathcal{A}$ on image luminance distribution by studying $L(x^{i}_j, p)$ as a function of $L(x^{pp}_j, p)$. To do so, we assume  as a functional form for Eq.\ref{eq:lum_operator}:
\begin{equation}
	L(x^{pp}_j, p) = \frac{L(x^{i}_j, p)^{\gamma}}{L(x^{i}_j, p)^{\gamma} + \mu^{\gamma}}
	\label{eq:lum_formula}
\end{equation}
inspired by the work of Naka-Rushton \cite{Naka1966} on the animal visual system. Eq.\ref{eq:lum_formula} represents a modified gamma compression \cite{Poynton2003}, with the additional global parameter $\mu$  that aims to capture the difference in tone mapping for images at different luminosity levels.

\begin{figure*}[!t]
	\centering
	\includegraphics[scale=0.25]{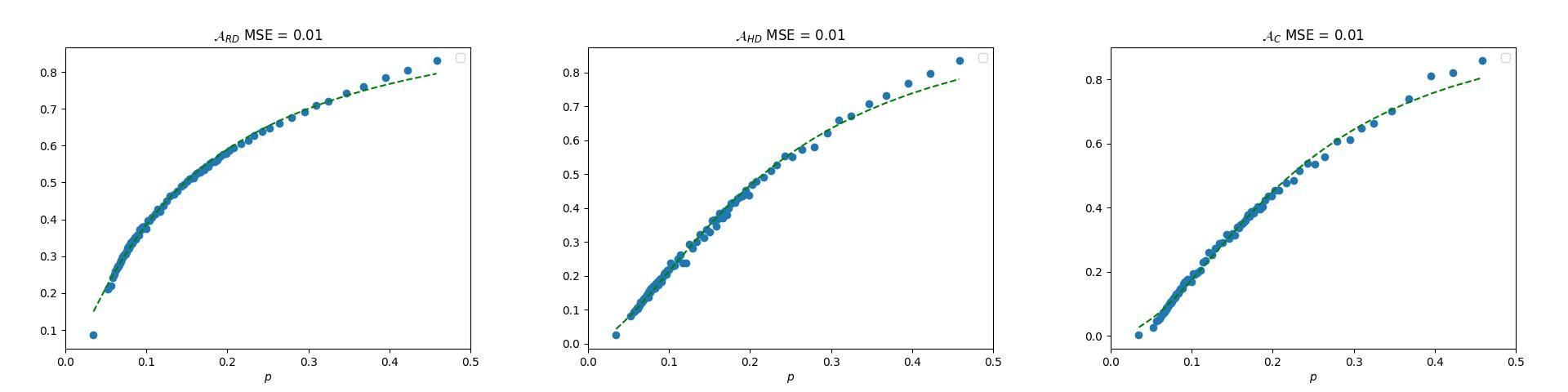}
	\caption{Examples of luminance distributions fits for classical (Reinhard), deep-learning (HDRNET) and manual (ExpertC MIT5K) tone mapping operator}\label{fig:fit_lum}  
	\label{fig:lum_group_hue_distribution}
\end{figure*}

To avoid overfitting, we randomly subsampled $P=100$ pixels to fit Eq.\ref{eq:lum_formula}. Fig. \ref{fig:fit_lum} shows that satisfactory fits can be obtained and we found the results presented here to be not too sensitive to the value of $P$.

Finally, to obtain a comprehensive view of the effect of $\mathcal{A}$ on $\mathcal{D}$, we fit Eq.\ref{eq:lum_formula} for each image pair, deriving a distribution of fit parameters $\left ( \gamma_j^{\mathcal{A}}, \mu_j^{\mathcal{A}} \right )$ on the full dataset.

Similarly, to quantify the effect of the denoising algorithm on the full dataset,  we computed for each image couple $\left (  x_j^i, x_j^{pp}  \right )$ the change of the local pixel variance due to the denoising algorithm
\begin{equation}
	\Delta \bar{\sigma}_j^{\mathcal{A}} =  \left|  \bar{\sigma} \left ( x_{pp}= \mathcal{A}(x_j) \right ) - \bar{\sigma}\left ( x_j \right ) \right|
\label{eq:delta_var}
\end{equation}

where $\bar{\sigma}$  is the average of the local variances computed on 16x16 windows for each image.

\begin{figure*}[!t]
	\centering
	\includegraphics[scale=0.4]{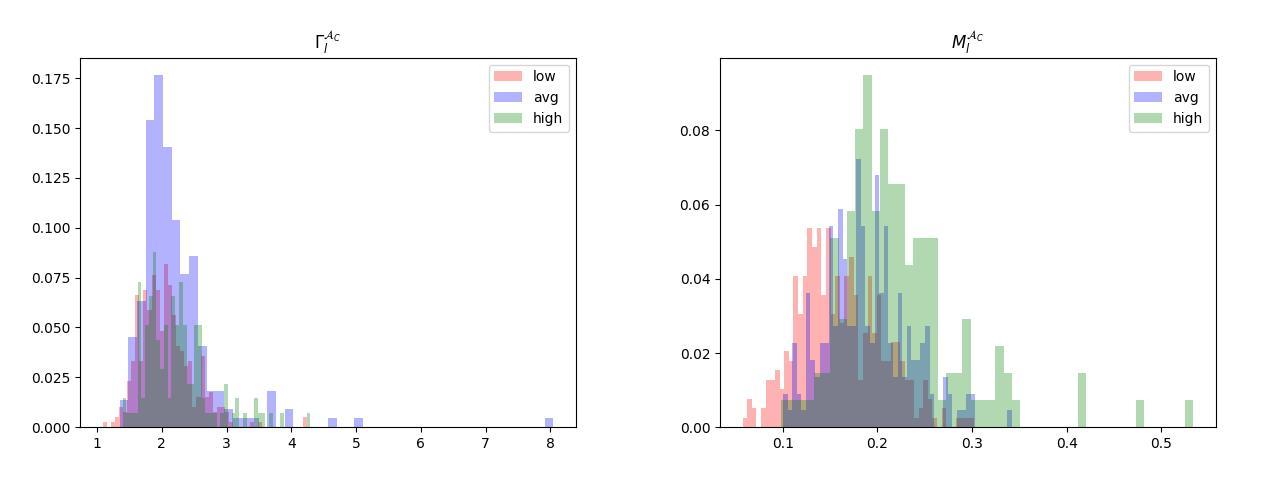}
	\caption{$\Gamma^{\mathcal{A}_{C}}_l$ and $M^{\mathcal{A}_{C}}_l$ distributions at different luminosity levels}   
	\label{fig:gamma_mu_distr_defin}  
\end{figure*}

\begin{figure*}[!t]
	\centering
	\includegraphics[width=0.5\linewidth]{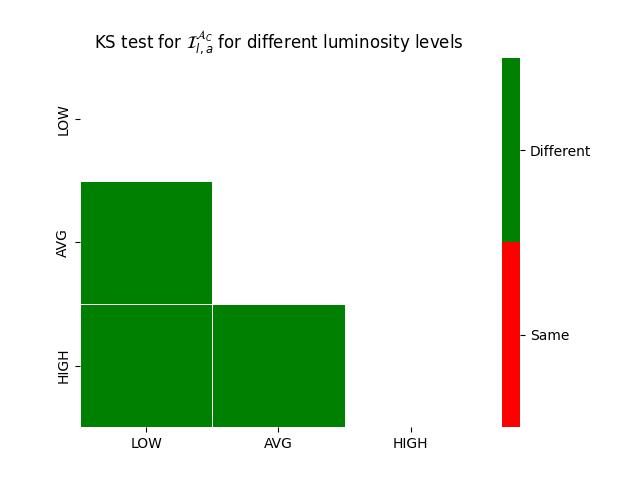}
	\caption{$\mathcal{I}_{l,a}^{\mathcal{A}_T}$  for images with different luminosity levels for $\mathcal{A}_C$ on $\mathcal{D}_{5K}$ }   
	\label{fig:ds_luminosity}
\end{figure*}

As discussed in Sec.\ref{sec:introduction}, one of the main goals of our analysis is to test the relevance of \clustname  for IE algorithms, i.e. to determine if images which belong to the same cluster have been enhanced similarly (and conversely for images in different clusters).

In Sec.\ref{subsection:unsuper_clustering} and \ref{subsection:quantify_algo} we showed how \clustname and  the fit of Eq.\ref{eq:lum_formula} associate to each image pair  $\left (x^{i}_j, x^{pp}_j  \right )$ a series of descriptors $\left ( d_{j,1}, \cdots, d_{j,k} \right )$, a cluster label $c^l_j$ and two fitting parameters  $\left ( \gamma_j^{\mathcal{A}}, \mu_j^{\mathcal{A}} \right )$. Combing these results, we can define the empirical distributions of fitting parameters for subsets of $\mathcal{D}$ as:
\begin{equation}
	\begin{split}
		\Gamma_l^{\mathcal{A}} = \left \{  \gamma_1^{\mathcal{A}}, \cdots, \gamma_{n_l}^{\mathcal{A}} \right \} \\
		 M_l^{\mathcal{A}} = \left \{ \mu_1^{\mathcal{A}}, \cdots, \mu_{n_l}^{\mathcal{A}}  \right \}
	\end{split}
\label{eq:gamma_mu_distr_defin}
\end{equation}
where $n_l$ is the number of images belonging to subset.

As a matter of example, in Fig. \ref{fig:gamma_mu_distr_defin} we show the $\Gamma_l^{\mathcal{A}}$ and $M_l^{\mathcal{A}}$ distributions at different luminosity levels for a tone mapping operator $\mathcal{A}$, namely ExpertC of MIT5K. By simple visual investigation, one can see how $\mathcal{A}$ applies different tone mapping corrections for images with different luminosity levels and, as expected, lower luminosity levels correspond to lower values of $\mu$.

Moreover, Eq.\ref{eq:gamma_mu_distr_defin} allows us to rephrase the cluster relevance question in more precise terms: asking if clusters are relevant for the operators $\mathcal{A}$ is in fact equivalent to asking for which combinations of cluster indexes $l$ and $a$ the following statements are true:
\begin{itemize}
\item is $\Gamma_l^{\mathcal{A}}$ statistically different to $\Gamma_a^{\mathcal{A}}$ if $l \neq  a$  ?
\item is $M_l^{\mathcal{A}}$ statistically different to $M_a^{\mathcal{A}}$ if $l \neq  a$  ?
\end{itemize}
and verifying if those combinations reflect the expected dynamics of $\mathcal{A}$ (e.g. images with different luminosity levels are in different clusters for tone mapping operators).

Both statements can be tested using a KS two-sided test (for which we choose a threshold for the $p-value$ of 0.05). Specifically, for a given $\mathcal{A}$, the test results can be effectively summarized using the cluster difference indicator:
\begin{equation}
\mathcal{I}^{\mathcal{A}_T}_{l,a} = \left\{\begin{matrix}
	1 & \mathrm{if} \, \mathrm{KS}\left ( \Gamma^{\mathcal{A}}_a , \Gamma^{\mathcal{A}}_l  \right ) < 0.05 \, \mathrm{or} \, \mathrm{KS}\left ( M^{\mathcal{A}}_a , M^{\mathcal{A}}_l  \right ) < 0.05, \\ 
	0 &  \mathrm{otherwise}
\end{matrix}\right.
\label{eq:cluster_indicator}
\end{equation} 
where $\mathrm{KS}\left ( \Gamma^{\mathcal{A}}_a , \Gamma^{\mathcal{A}}_l  \right )$  and $\mathrm{KS}\left ( M^{\mathcal{A}}_a , M^{\mathcal{A}}_l  \right )$ are the $p-values$ of KS tests performed on the $\Gamma^{\mathcal{A}}$ and $M^{\mathcal{A}}$ of the clusters that we are investigating.

If, for a  combination of clusters $(l,a)$ at least one of the distributions of parameters $(\Gamma, M)$ is statistically different, $\mathcal{I}^{\mathcal{A}}_{l,a} = 1 $ signals that images in clusters $l$ and  $a$ are subject to statistically distinct IE transformations.

As a matter of example, Fig.\ref{fig:ds_luminosity} shows the results of the K-S tests for $\Gamma_l$ and $M_l$ at varying level of luminosity for ExpertC of MIT5K, for which we display in green $\mathcal{I}^{\mathcal{A}}_{l,a} = 1$ and red otherwise.  For this particular operator, we can see that $\mathcal{I}^{\mathcal{A}}_{l,a} = 1$ for each combination of luminosity levels. This quantitatively confirms that images with different luminosity levels are treated differently by the ExpertC and clustering images according to their luminosity is a reasonable choice for this tone mapping operator.

In a similar fashion, we can build the cluster difference indicator for the denoising algorithm $\mathcal{I}^{\mathcal{A}_{D}}_{l,a} $ as 

\begin{equation}
	\mathcal{I}^{\mathcal{A}_{D}}_{l,a} = \left\{\begin{matrix}
		1 & \mathrm{if} \, \mathrm{KS}\left ( \Sigma^{\mathcal{A}}_a , \Sigma^{\mathcal{A}}_l  \right ) < 0.05  \\ 
		0 &  \mathrm{otherwise}
	\end{matrix}\right.
	\label{eq:cluster_indicator_var}
\end{equation} 

where $\Sigma^{\mathcal{A}}_{a,l}$ are the distributions of the variation of the local variances $\Delta \bar{\sigma}_j^{\mathcal{A}}$ as defined by Eq. \ref{eq:delta_var}, which we will use below to quantify the impact of denoising algorithms.

\section{Experimental Results on Clustering}
\label{sec:results}
The main goal of this section is to test the relevance of clusters obtained by \clustname, especially when compared to those obtained using semantic  information. In particular, we will test two alternative clustering methodologies:
\begin{itemize}

\item[--] {an approach leveraging user-defined descriptors from the MIT5K dataset, which provides 4 manually annotated semantic descriptors for each image pair. The categories of descriptors are location, time, subject, light. We discarded the column with the light descriptor, as it was highly correlated with time and location. Using these descriptors, we follow the procedure in}  Sec.\ref{subsection:unsuper_clustering} {to create clusters.} 
\item[--] {an alternative approach based on pretrained ResNet50-places365 network} \cite{zhou2017places}  {as cluster labels: these attributes describe interesting proprieties of the images (e.g. indoor/outdoor); however, there is no automated mechanism to identify which proprieties are most relevant. We focus here on only 4 attributes, namely natural light (Nat), no horizon (NH), man-made (Man), open area (Op), which we selected among the almost 200 original descriptors because they were the most present in the dataset. Again, we follow the procedure in Sec.}\ref{subsection:unsuper_clustering}  {to create clusters.}
\end{itemize}

 {While not exhaustive, we believe that these models represent key examples of both traditional and neural network-based semantic clustering algorithms. Moreover, the testing method described in Sec.}\ref{subsection:quantify_algo} {can easily be extended to other clustering algorithms, allowing an easy comparison of} \clustname.

For what regards the IE algorithms, this work focuses on  HDRNET \cite{gharbi2017deep} for tone mapping and NAFNet \cite{chen2022simplebaselinesimagerestoration} for denoising since they are SOTA and open access implementation exist.

Finally, in our analysis we fix \clustname's hyperparameter as $\mathcal{C} = \left ( 0, \frac{\pi}{3}, \frac{2 \pi}{3},  \frac{4 \pi}{3},  \frac{5 \pi}{3} \right )$ , i.e.  (red, yellow, green, blue, magenta),  $\Delta \theta =  \frac{\pi}{6} $ and $K=100$. We found that for these values the performances of \clustname are both stable and satisfactory for the datasets we analyzed.

\subsection{Dataset}
\label{subsection:datasets}

Two main datasets have been used to carry out the analysis presented here:
\begin{itemize}
	\item[--]  $\mathcal{D}_{5K}$: 5000 image pairs (input and ExpertC) from MIT5K dataset (ExpertC) \cite{fivek}, extensively used in literature for global tone mapping tasks
	\item[--]  $\mathcal{D}_{noisy}$: a synthetic, noisy dataset  created by adding synthetic noise to 5000 image pairs ExpertC subset of $\mathcal{D}_{5K}$. 
\end{itemize}

More precisely, we constructed $\mathcal{D}_{noisy}$ by modifying the pixel values of ground truth images $ x^{gt}_j$ of $\mathcal{D}_{5K}$ as
\begin{equation}
	x^{noisy}_j \sim \mathcal{N} \left ( x^{gt}_j, a_j  x^{gt}_j + b_j \right )
	\label{eq:noise}
\end{equation}

 {where $a_j$ and $b_j$ are free parameters independently sampled from a uniform distribution for each image, with bounds $a_j \in \left [ 0.3 , 0.5 \right ]$ and $b_j \in \left [ 0.1 , 0.2 \right ]$. This follows the methodology proposed by Foi et al.} \cite{Foi2008}  {, ensuring a realistic noise distribution suitable for image enhancement (IE) tasks.}

To evaluate the effectiveness of \clustname in enhancing performance on underrepresented classes, we deliberately increased the noise in the two least represented classes of $\mathcal{D}_{5K}$ by drawing $a_j, b_j$ from a uniform distribution with a higher mean than that used for the remainder of the dataset. {More precisely,  we fix $a_j \in \left [ 0.6 , 1.0 \right ]$ and $b_j \in \left [ 0.2 , 0.4 \right ]$ on unrepresented classes, to introduce approximately twice the noise level relative to the rest of the dataset.}

\begin{table}[!t]
	\renewcommand{\arraystretch}{1.3}
	\caption{Number of clusters $\# \, c_l$, the silhouette score \silscore and percentage of noise points $\% \, c_s$ for the clustering algorithms studied here.}
	\centering
	\begin{tabular}{c||c|c|c}
		\hline
		&  $\# \, c_l$ & \silscore  & $\% \, c_s$\\
		\hline\hline
		\clustname $({D}_{5K})$ & $12$ & $0.67$  & $0.00$\\
		\hline
		 MIT5K $({D}_{5K})$ & $17$ & $0.90$  & $0.52$ \\
		\hline
		ResNet Places365 $({D}_{5K})$ & $5$ & $0.87$ &  $1.86$\\
		\hline
	\end{tabular}
	\label{table:clusters}
\end{table}

\subsection{Clusters' relevance}
\label{subsection:clu_rel}
\begin{figure*}[!t]
	\centering
	\includegraphics[scale=0.25]{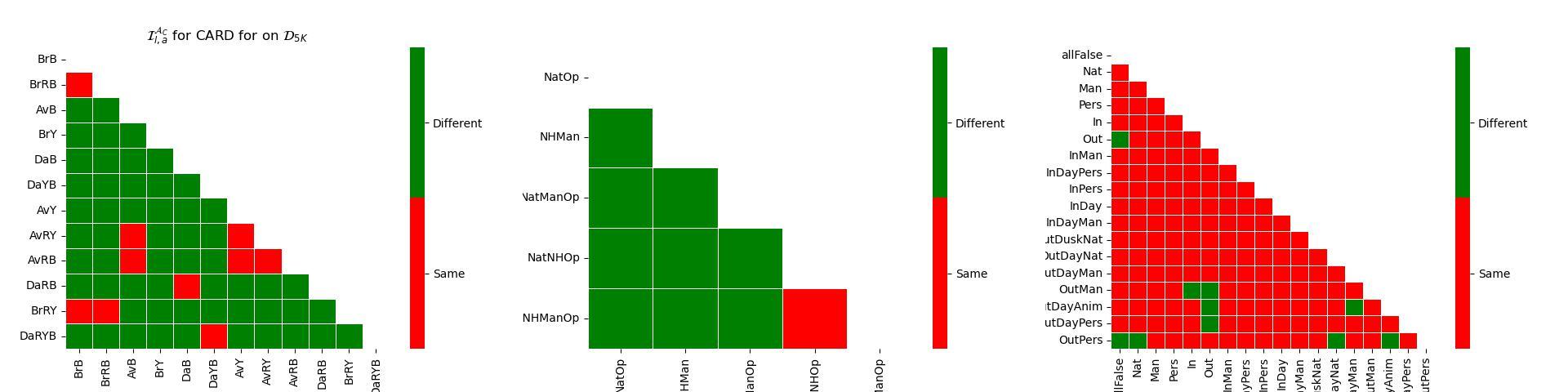}
	\caption{$\mathcal{I}^T_{l,a}$ for clusters obtained by ResNet50 (left) and by MIT5K (right) for $\mathcal{D}_{5K}$}
	\label{fig:ds_test_clustname}
\end{figure*}

Fig.\ref{fig:ds_test_clustname} shows the results of K-S tests obtained by applying \clustname on $\mathcal{D}_{5K}$, using the indicator cluster difference $\mathcal{I}_{l,a}^{\mathcal{A}}$. Firstly, we can notice that clusters defined by \clustname for  $\mathcal{D}_{5K}$ are statistically different only if they have different luminosity levels, as expected for a global tone mapping operator such as HDRNET. This shows how clusters obtained by \clustname capture correctly the dynamics of IE operators and  partition the datasets in a way that can be exploited for large scale image dataset analysis (e.g. spotting unrepresented ``classes" in the dataset, see Section \ref{sec:results_resampling}).

These results are even more convincing when comparing them to those for the clusters obtained using ResNet Places365 and manual MIT5K labels.
Specifically, for the latter, $\mathcal{I}^T_{l,a}=0$ for almost any $l,a$,  effectively resulting in a random clustering with respect to this task.  On the other hand,  clusters generated using ResNet Places365 exhibit some degree of sensitivity to the dynamics of a global tone mapping operator, but less accurately than \clustname, likely due to the presence of images with varying luminosity levels within the clusters (Table \ref{table:scenes_cluster_luminosity_mit}).

\begin{table}[ht]
    \centering
    \begin{minipage}{.5\textwidth}
        \centering
		\caption{\clustname's clusters statistics for $\mathcal{D}_{5K}$}
        \begin{tabular}{c||c|c|c}
			\hline
			$c_l$ & labels  & $\% \, c_l$\\
			\hline\hline
			DaRB & \{dark lum, red, blue\}  & $10.27$ \\
			\hline
			DaRYB & \{dark lum, red, yellow, blue\}  & $8.32$ \\
			\hline
			DaB & \{dark lum, blue\}  & $19.6$ \\
			\hline		
			DaYB & \{dark lum, yellow, blue\}  & $15.63$ \\
			\hline
			AvB & \{average lum, blue\}  & $9.95$ \\
			\hline
			AvRB & \{average lum, red, blue\}  & $5.57$ \\
			\hline
			AvRY & \{average lum, red, yellow\}  & $5.49$ \\
			\hline
			AvY & \{average lum, yellow\}  & $12.09$ \\
			\hline
			BrB & \{bright lum, blue\}  & $2.88$ \\
			\hline
			BrRB & \{bright lum, red, blue\}  & $2.61$ \\
			\hline
			BrRY & \{bright lum, red, yellow\}  & $2.48$ \\
			\hline
			BrY & \{bright lum, yellow\}  & $5.15$ \\
			\hline
		\end{tabular}
		\label{clusters_tm}
    \end{minipage}%
    \begin{minipage}{.5\textwidth}
        \centering
		\caption{\clustname's clusters statistics for $\mathcal{D}_{noisy}$}
        \begin{tabular}{c||c|c|c}
			\hline
			$c_l$ & labels  & $\% \, c_l$\\
			\hline\hline
			AvB & \{average lum, blue\}  & $6.84$ \\
			\hline
			BrB & \{bright lum, blue\}  & $6.88$ \\
			\hline
			DaB & \{dark lum, blue\}  & $17.34$ \\
			\hline		
			DaY & \{dark lum, yellow\}  & $24.26$ \\
			\hline
			AvY & \{average lum, yellow\}  & $19.52$ \\
			\hline
			BrY & \{bright lum, yellow\}  & $25.16$ \\
			\hline
		\end{tabular}
		\label{clusters_dn}
		\vspace{90pt}
    \end{minipage}
\label{table:cluster_stats}
\end{table}

\begin{table}[!t]
	\renewcommand{\arraystretch}{1.3}
	\caption{ResNet50-places365's clusters luminosity levels as defined by \clustname for $\mathcal{D}_{5K}$}
	\centering
	\begin{tabular}{c||c|c|c}
		\hline
		& $\%$ Low Lum & $\%$ Avg Lum & $\%$ High Lum\\
		\hline\hline
		NHMan & $0.66$ & $0.27$  & $0.07$\\
		\hline
		NatManOP & $0.35$ & $0.41$  & $0.23$\\
		\hline
		NatNHManOP & $0.51$ & $0.35$  & $0.14$\\
		\hline
		NatNHOp  & $0.61$ & $0.32$  & $0.07$\\
		\hline
		NatOp  & $0.34$ & $0.41$  & $0.25$\\
		\hline
	\end{tabular}
	\label{table:scenes_cluster_luminosity_mit}
\end{table}

\begin{figure*}[!t]
	\centering
	\includegraphics[scale=0.25]{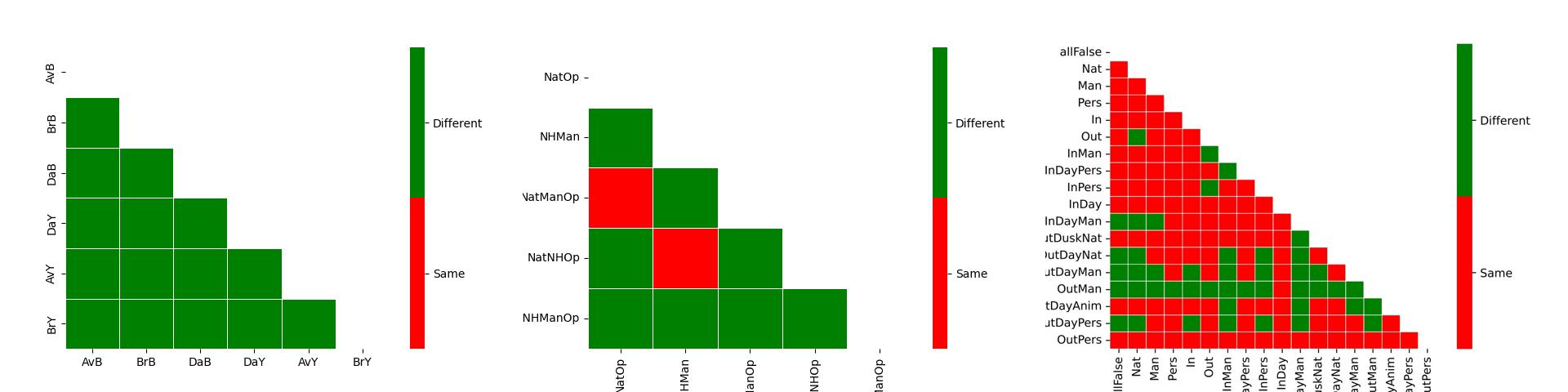}
	\caption{$\mathcal{I}_{l,a}$ for clusters obtained by ResNet50 (left) and by MIT5K (right) for $\mathcal{D}_{noisy}$}
	\label{fig:ds_test_clustname_denoising}
\end{figure*}

\section{Experimental Results on Resampling}
\label{sec:results_resampling}
In this section, we present the experimental results obtained by applying the oversampling  to the two Image Enhancement tasks: HDRNet for Tone Mapping and  NAFNet for Denoising.

{All results presented here have been obtained by training HDRNET and NAFNet on a 1 GPU machine with 11 GB of memory and 4352 cores for $30$ epochs. We used the PSNR as training and validation metric and ADAM optimizer with learning rate of $10^{-4}$. Finally, we kept as best checkpoint the one with the lowest validation loss during the training.}

To establish baseline results, we performed a 75-25 train-test split on both $\mathcal{D}_{5K}$ and $\mathcal{D}_{noisy}$ and trained HDRNET and NAFNet on the original datasets. Subsequently, we generated resampled datasets by oversampling the minority classes within both \clustname and ResNet Places365 clusters with a replication factor of $n\_overs=3$ times.

Given that the results in Section \ref{subsection:clu_rel} indicated that clusters based on MIT-5K manual labels were effectively random, we opted against their direct use. Instead, we employed a dataset where we randomly oversampled the entire original dataset to match the number of images oversampled from \clustname's minority classes.

\begin{table}[!t]
	\renewcommand{\arraystretch}{1.3}
	\caption{Difference between PSNR obtained using the resampling techniques and baseline on full datasets}
	\centering
	\begin{tabular}{c||c|c|c}
	Task  & \makecell{$\Delta$ PSNR (dB) \\ Random }  & \makecell{$\Delta$ PSNR (dB) \\Resampled ResNet } & \makecell{$\Delta$ PSNR (dB) \\ Resampled \clustname}\\
	\hline \hline
	Tone Mapping  & 0.00 & 0.08 & \textbf{0.22} \\
	Denoising  & -1.5 &   -0.09 & \textbf{1.18} \\
	\end{tabular}
	\label{table:results_resampling}
	\end{table}
	
\begin{table}[!t]
	\renewcommand{\arraystretch}{1.3}
	\caption{Difference between PSNR obtained using the resampling techniques and baseline on minority classes}
	\centering
	\begin{tabular}{c||c|c|c}
		Task  & \makecell{$\Delta$ PSNR (dB) \\ Random}  & \makecell{$\Delta$ PSNR (dB) \\Resampled ResNet} & \makecell{$\Delta$ PSNR (dB) \\ Resampled \clustname}\\
		\hline \hline
		Tone Mapping  & 0.01 & 0.12 & \textbf{0.60} \\
		Denoising  & -0.28 &  0.35  & \textbf{2.73} \\
	\end{tabular}
	\label{table:results_resampling_minority}
\end{table}

Tables \ref{table:results_resampling} and \ref{table:results_resampling_minority} show the performance changes (in PSNR) while using a random, ResNet and \clustname's clusters as a basis for resampling. We can see that, as expected from the results of Sec.\ref{subsection:clu_rel}, \clustname outperforms the other methods both on the full dataset and on the minority (oversampled) classes.

\begin{figure*}[!t]
	\centering
	\includegraphics[scale=0.25]{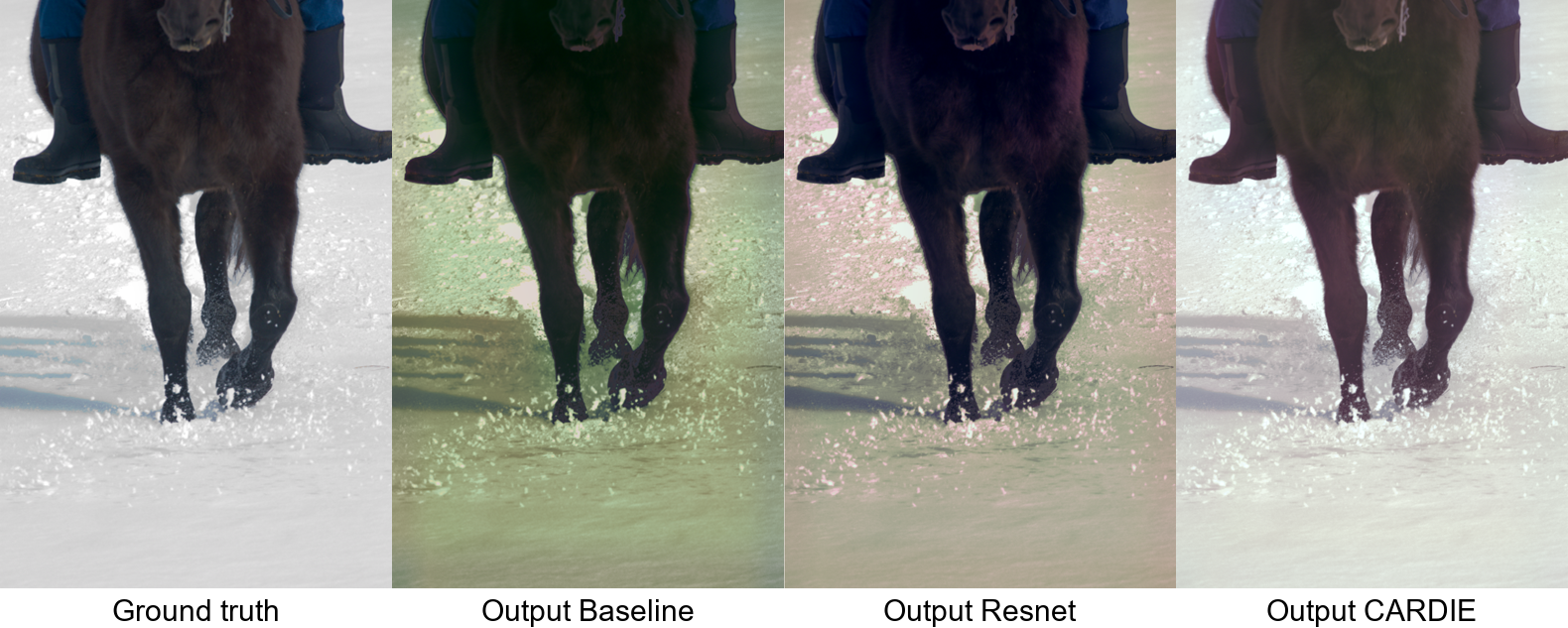}
	\caption{Example of visual results for the Tone Mapping task using HDRNet and $\mathcal{D}_{5K}$}
	\label{fig:tone_mapping_sample}
\end{figure*}

\begin{figure*}[!t]
	\centering
	\includegraphics[scale=0.08]{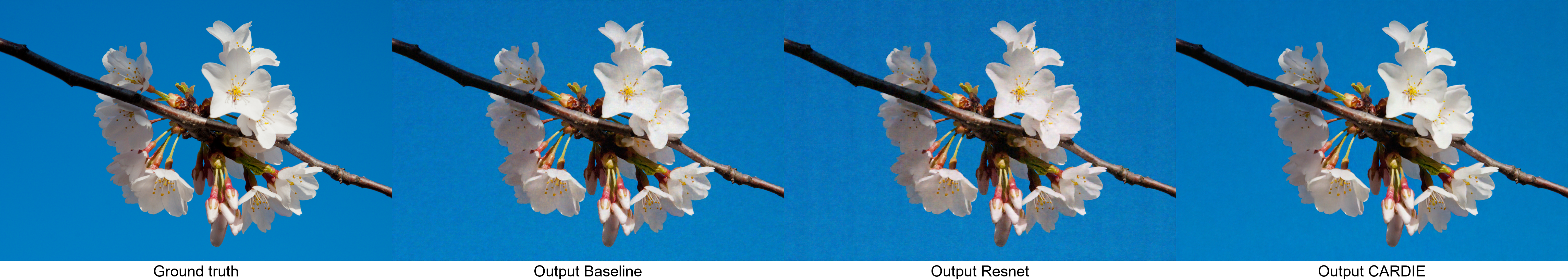}
	\caption{Example of visual results for the Denoising task using NAFNet and $\mathcal{D}_{noisy}$}
	\label{fig:denoising_sample}
\end{figure*}

Figures \ref{fig:tone_mapping_sample} and \ref{fig:denoising_sample} respectively illustrate the results obtained using the three clustering algorithms for the Tone mapping and Denoising tasks. These images visually confirm the superior performance of \clustname compared to the random and ResNet ones.

These results confirm that \clustname captures more accurately the underlying dynamics of IE algorithms compared to semantic approaches and can be used to enhance their performance, without the need of modifying the algorithm architectures.

\section{Conclusions}
\label{sec:Conclusion}

We introduced a new two-step image clustering algorithm \clustname which automatically labels images, according to their luminosity and color contents and then clusters them with HDBSCAN.

We also introduced a method for evaluating cluster relevance, based on changes in post-processed image luminance and local variance, with respect to two deep learning-based IE operators: a tone mapper (HDRNET) and a denoiser (NAFNet).

Thanks to this method, we proved that clusters obtained by \clustname are more relevant for these IE tasks than the ones obtained by the semantic clustering methods. Specifically, we statistically demonstrated that images within the same \clustname clusters tend to undergo similar image enhancement (IE) processes, especially when compared to those in different clusters.

Finally, we demonstrated that the clusters generated by \clustname can serve as a foundation for resampling IE datasets, significantly enhancing the performance of CNN-based tone mapping and denoising algorithms.

For these reasons, we consider \clustname a valuable tool for large-scale image dataset analysis in the domain of image enhancement, and have therefore made it publicly available on our \href{https://github.com/GiuliaBonino/CARDIE}{github}.

\section*{Discloures}
\label{app:disc}
 
All authors declare no conflicts of interest in this study.

\section*{Code-Data Availability Statement}
\label{app:code availabiliy}

To foster the usage and ensure the reproducibility of the analysis conducted here, we release \clustname on our \href{https://github.com/GiuliaBonino/CARDIE}{github}.

\section*{Large Language Model Statement}
\label{app:llms}

Large Language Models, in particular \href{https://gemini.google.com/}{Gemini}, were used for minor syntax and spelling review.

\newpage

\appendix
\section{Clusters' samples for \clustname}
\label{app:cluster_sample}

\begin{figure}[!t]
	\centering
	\includegraphics[width=\linewidth]{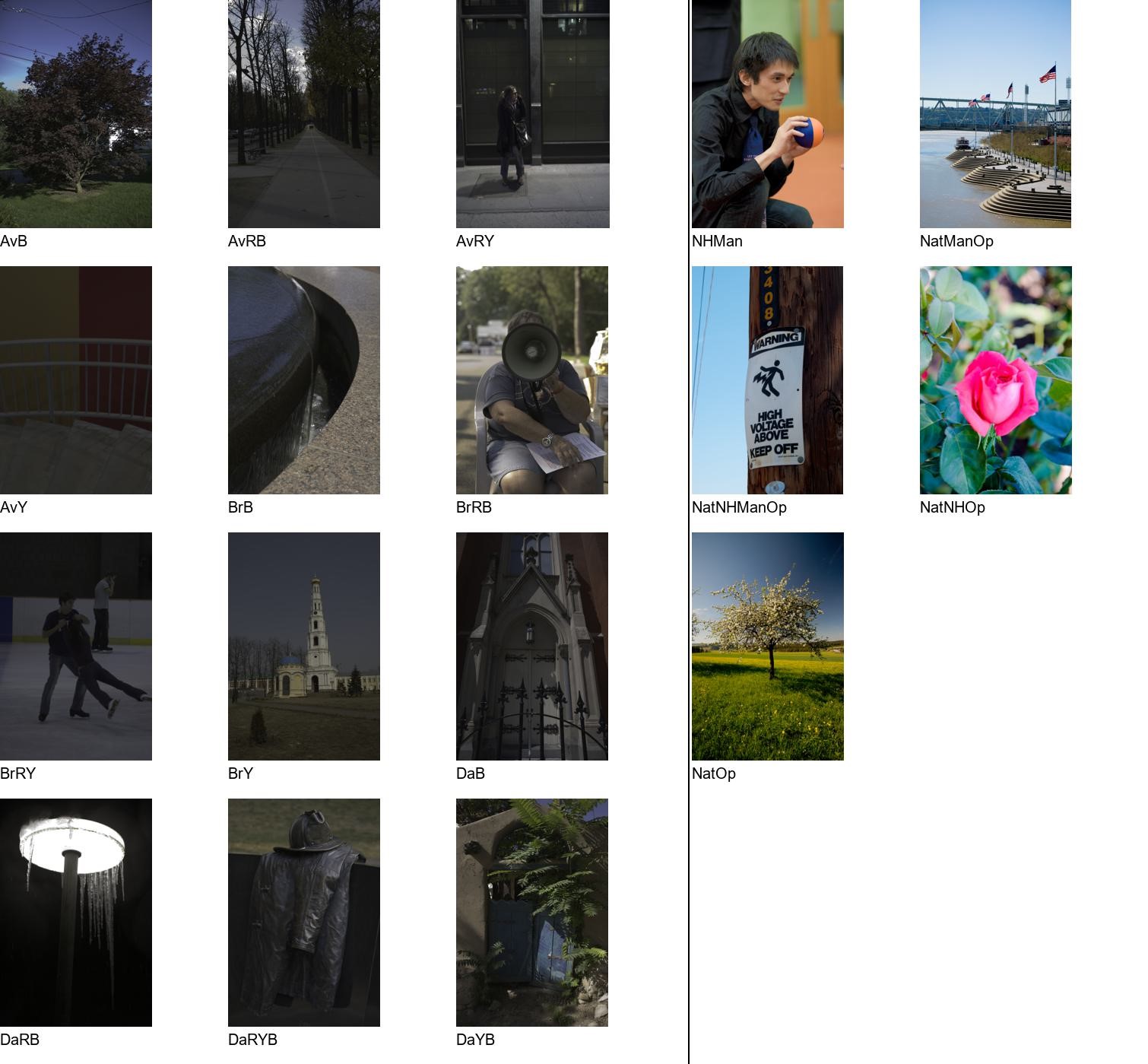}
	
	\caption{Sample of images clustered by \clustname for $\mathcal{D}_{5K}$ (left) ResNet50-places365  for $\mathcal{D}_{5K}$ (right, ground truth images are displayed for clarity)}
	\label{fig:clusters_clustname}
\end{figure}

Fig.\ref{fig:clusters_clustname}  show samples of images belonging to different clusters obtained by applying respectively \clustname and  ResNet50-places365 on ${D}_{5K}$. This visually confirms that both clustering algorithms correctly classify images according to their descriptors, color-luminosity for \clustname and semantic for ResNet50-places365.

\section{Results on HDR+ dataset}
\label{app:hdrplus}

{In this section, we provide an additional comparison between} \clustname {and ResNet Places 365, conducted on a subset of 1,684 image pairs from the HDR+ dataset } \cite{hasinoff2016burst}  {which was designed to support research in computational photography, particularly for HDR  and low-light images.}

\begin{figure*}[!t]
	\centering
	\includegraphics[scale=0.4]{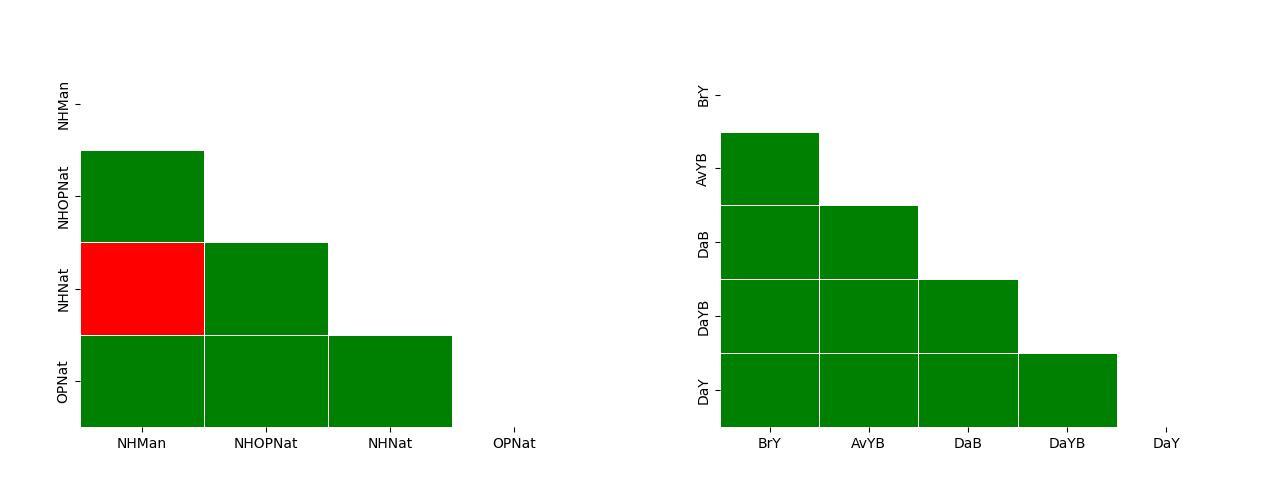}
	\caption{$\mathcal{I}^T_{l,a}$ for clusters obtained by ResNet50 (left) and by \clustname (right) for HDR+}
	\label{fig:ds_test_clustname_hdrplus}
\end{figure*}

Fig. \ref{fig:ds_test_clustname_hdrplus}  {shows the $\mathcal{I}^T_{l,a}$ indicator for clusters by ResNet50 (left) and by } \clustname {(right) on HDR+, obtained with the same procedures outlined in} Sec.\ref{section:method}   {. These results confirm that} \clustname  {captures the dynamics of IE operators more precisely than ResNet50, in particular since each cluster corresponds to a distinct combinations of luminosity levels and colors,  treated differently by the HDR+ image pipeline.}

 {This does not apply to the clusters produced by ResNet50 which, as shown by Table } \ref{table:scenes_cluster_luminosity_hdrplus},  { do not group together images with similar colors and/or luminosity levels. In particular,  the $NHNat$ and $NHOPNat$ clusters exhibit similar combinations of dominant colors and luminosity levels. Consequently, the HDR+ pipeline processes them in an indistinguishable manner, rendering these clusters useless for this task.}

 {Since this confirms the results found in Sec.} \ref{sec:results} {and being the HRD+  mostly a classical pipeline, we did not repeat the oversampling experiments presented in} \ref{sec:results_resampling}  {on the HRD+ dataset.}

\begin{figure}[!t]
	\centering
	\includegraphics[width=5.5in]{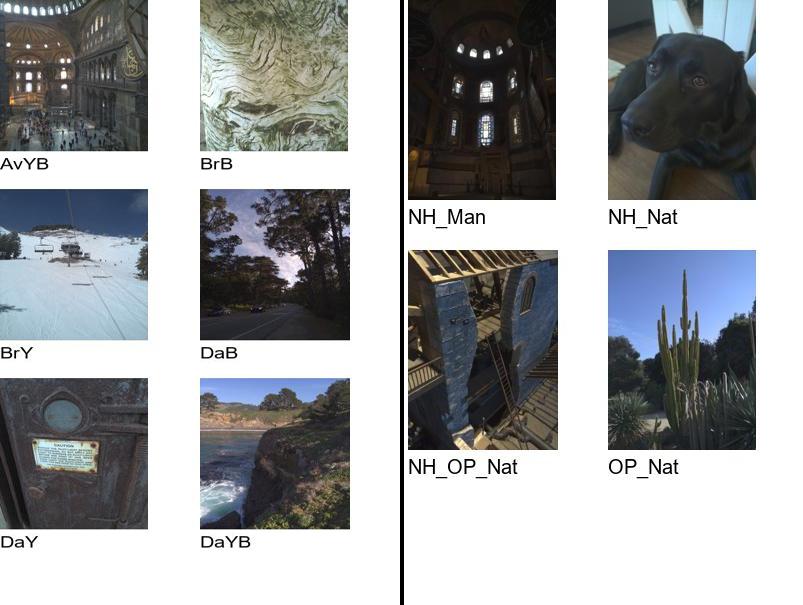}
	\caption{Sample of images clustered by \clustname for $\mathcal{D}_{5K}$ (left) ResNet50-places365 for HDRPLUS (ground thruth images are displayed for clarity) }
	\label{fig:clusters_places365_HDRPLUS}
\end{figure}

\begin{table}[!t]
	\renewcommand{\arraystretch}{1.3}
	\caption{ResNet50-places365's clusters luminosity levels and colors percentages as defined by \clustname for HDR+ }
	\centering
	\begin{tabular}{c||c|c|c|c|c|c}
		\hline
		& $\%$ Low Lum & $\%$ Avg Lum & $\%$ High Lum & $\%$ B  & $\%$ Y & $\%$ YB \\ 
		\hline\hline
		NHOPNat & $0.59$ & $0.23$  & $0.17$  & $0.23$  & $0.31$  & $0.46$\\
		\hline
		NHNat & $0.77$ & $0.10$  & $0.13$  & $0.41$  & $0.11$  & $0.48$\\
		\hline
		NHMan  & $0.81$ & $0.10$  & $0.09$  & $0.66$  & $0.04$  & $0.30$\\
		\hline
		OPNat & $0.26$ & $0.31$  & $0.41$  & $0.06$  & $0.54$  & $0.40$\\
		\hline
	\end{tabular}
	\label{table:scenes_cluster_luminosity_hdrplus}
\end{table}

\section{$L$ and $\mathcal{H}$ distributions}
\label{app:lum_hue_distr}

Throughout all the analysis conducted in this paper we computed the luminance and hue distributions for each image in terms of its RGB content.

In particular, the luminance distribution for image $x_j$ was computed as:
\begin{equation}
	L(x_j, p)  = 0.299 R[x_j, p] + 0.587  G[x_j, p] + 0.114  B[x_j, p]
	\label{eq:lum_rgb}
\end{equation} 
where $p$ is the pixel index.

On the other side, the hue distribution is computed from the opponent space coordinates $O_1$ and $O_2$:
\begin{equation}
		\begin{split}
			O_1[x_j, p] = \left ( R[x_j, p] + G[x_j, p] + B[x_j, p] - 1.5 \right ) / 1.5 \\
			O_2[x_j, p] = \left ( R[x_j, p] - G[x_j, p] \right )
		\end{split}
	\label{eq:opp_space_def}
\end{equation}
as: 
\begin{equation}
	\mathcal{H}(x_j, \theta) = \arctan(O_2[x_j, p], O_1[x_j, p])
	\label{eq:hue_def}
\end{equation}
where $\theta$ is then shifted in the interval $\left ( 0, 2 \pi \right )$ for convenience. 
 

\newpage

\vspace{2ex}\noindent\textbf{Giulia Bonino} received her Master’s degree in Data Science and Engineering through a double degree program between Politecnico di Torino and EURECOM. She completed her Master’s thesis at the Huawei Nice Research Center, where she focused on computer vision.

\vspace{2ex}\noindent\textbf{Luca Alberto Rizzo}  received his Ph.D. in Physics from Université Paris-Saclay in 2017. He is currently a Researcher at the Huawei Nice Research Center, where his work focuses on computer vision, with particular emphasis on interpretability, efficiency, and automatic image clustering.


\bibliography{report}   

\begin{thebibliography}{10}

\bibitem{chen2022simplebaselinesimagerestoration}
L.~Chen, X.~Chu, X.~Zhang, {\em et~al.}, ``Simple baselines for image
  restoration,''  (2022).

\bibitem{Zhao2022}
L.~Zhao, A.~Abdelhamed, and M.~Brown, ``Learning tone curves for local image
  enhancement,'' {\em IEEE Access} {\bf 10}, 1--1  (2022).

\bibitem{GharbiCBHD17}
M.~Gharbi, J.~Chen, J.~T. Barron, {\em et~al.}, ``Deep bilateral learning for
  real-time image enhancement,'' {\em CoRR} {\bf abs/1707.02880}  (2017).

\bibitem{hashmani2019accuracy}
M.~A. Hashmani, S.~M. Jameel, H.~Alhussain, {\em et~al.}, ``Accuracy
  performance degradation in image classification models due to concept
  drift,'' {\em International Journal of Advanced Computer Science and
  Applications} {\bf 10}(5)  (2019).

\bibitem{Khan2023}
A.~Khan and N.~Malim, ``Comparative studies on resampling techniques in machine
  learning and deep learning models for drug-target interaction prediction,''
  {\em Molecules (Basel, Switzerland)} {\bf 28}  (2023).

\bibitem{zhou2017places}
B.~Zhou, A.~Lapedriza, A.~Khosla, {\em et~al.}, ``Places: A 10 million image
  database for scene recognition,'' {\em IEEE Transactions on Pattern Analysis
  and Machine Intelligence}   (2017).

\bibitem{zhou2018semanticunderstandingscenesade20k}
B.~Zhou, H.~Zhao, X.~Puig, {\em et~al.}, ``Semantic understanding of scenes
  through the ade20k dataset,''  (2018).

\bibitem{lin2015microsoftcococommonobjects}
T.-Y. Lin, M.~Maire, S.~Belongie, {\em et~al.}, ``Microsoft coco: Common
  objects in context,''  (2015).

\bibitem{5539970}
J.~Xiao, J.~Hays, K.~A. Ehinger, {\em et~al.}, ``Sun database: Large-scale
  scene recognition from abbey to zoo,'' in {\em 2010 IEEE Computer Society
  Conference on Computer Vision and Pattern Recognition},  3485--3492  (2010).

\bibitem{VanGansbeke2020}
W.~V. Gansbeke, S.~Vandenhende, S.~Georgoulis, {\em et~al.}, ``Learning to
  classify images without labels,'' {\em CoRR} {\bf abs/2005.12320}  (2020).

\bibitem{caron2021unsupervisedlearningvisualfeatures}
M.~Caron, I.~Misra, J.~Mairal, {\em et~al.}, ``Unsupervised learning of visual
  features by contrasting cluster assignments,''  (2021).

\bibitem{Niu_2022}
C.~Niu, H.~Shan, and G.~Wang, ``Spice: Semantic pseudo-labeling for image
  clustering,'' {\em IEEE Transactions on Image Processing} {\bf 31},
  7264–7278  (2022).

\bibitem{CampelloHDBSCAN}
R.~J. G.~B. Campello, D.~Moulavi, A.~Zimek, {\em et~al.}, ``Hierarchical
  density estimates for data clustering, visualization, and outlier
  detection,'' {\em ACM Trans. Knowl. Discov. Data} {\bf 10}  (2015).

\bibitem{baligodugula2025unsupervisedlearningcomparativeanalysis}
V.~V. Baligodugula and F.~Amsaad, ``Unsupervised learning: Comparative analysis
  of clustering techniques on high-dimensional data,''  (2025).

\bibitem{ROUSSEEUW198753}
P.~J. Rousseeuw, ``Silhouettes: A graphical aid to the interpretation and
  validation of cluster analysis,'' {\em Journal of Computational and Applied
  Mathematics} {\bf 20}, 53--65  (1987).

\bibitem{Naka1966}
R.~W. Naka~KI, ``S-potentials from luminosity units in the retina of fish
  (cyprinidae),'' {\em J Physiol} {\bf 3}, 587--99  (1966).

\bibitem{Poynton2003}
C.~Poynton, {\em Digital Video and HDTV Algorithms and Interfaces}, Morgan
  Kaufmann Publishers Inc., San Francisco, CA, USA, 1~ed.  (2003).

\bibitem{gharbi2017deep}
M.~Gharbi, J.~Chen, J.~T. Barron, {\em et~al.}, ``Deep bilateral learning for
  real-time image enhancement,'' {\em ACM Transactions on Graphics (TOG)} {\bf
  36}(4), 118  (2017).

\bibitem{fivek}
V.~Bychkovsky, S.~Paris, E.~Chan, {\em et~al.}, ``Learning photographic global
  tonal adjustment with a database of input / output image pairs,'' in {\em The
  Twenty-Fourth IEEE Conference on Computer Vision and Pattern Recognition},
  (2011).

\bibitem{Foi2008}
A.~Foi, M.~Trimeche, V.~Katkovnik, {\em et~al.}, ``Practical
  poissonian-gaussian noise modeling and fitting for single-image raw-data,''
  {\em IEEE Transactions on Image Processing} {\bf 17}(10), 1737--1754  (2008).

\bibitem{hasinoff2016burst}
S.~W. Hasinoff, D.~Sharlet, R.~Geiss, {\em et~al.}, ``Burst photography for
  high dynamic range and low-light imaging on mobile cameras,'' {\em ACM
  Transactions on Graphics (Proc. SIGGRAPH Asia)} {\bf 35}(6)  (2016).

\end{thebibliography}
\bibliographystyle{spiejour}   


\listoffigures
\listoftables

\end{spacing}
\end{document}